\documentclass[letterpaper,10pt,journal,twoside]{IEEEtran}

  \usepackage{amsmath,amssymb,amsfonts,bm}
  \def\tymes{{\times}}
  \usepackage[
    free-standing-units = true,
    per-mode = single-symbol,
    product-units = power,
    product-symbol = \tymes,
    range-units = single,
    list-units = single,
    list-final-separator = {, and },
    uncertainty-mode = separate,
    bracket-ambiguous-numbers = false
  ]{siunitx}
  \usepackage{graphicx}
  \usepackage{xcolor}
  \usepackage[most]{tcolorbox}
  \usepackage{booktabs}
  \usepackage{balance}
  \usepackage{blindtext}
  \newif\ifhyperref{}\hyperreftrue{}
  \ifhyperref{}
    \usepackage[hidelinks,bookmarksopen=true]{hyperref}
    \usepackage[all]{hypcap}
  \else
    \DeclareRobustCommand{\href}[2]{#1}
    \DeclareRobustCommand{\nolinkurl}[1]{#1}
  \fi
  \usepackage{cite}
  \usepackage[nameinlink,capitalise]{cleveref}
  \usepackage{adjustbox}
  \usepackage{multirow}
  \usepackage{threeparttable}

  \usepackage{tikz}
  \usetikzlibrary{positioning, shapes.geometric, arrows.meta, chains, matrix, decorations.pathreplacing}

  \usepackage{pgfplots}
  \DeclareUnicodeCharacter{2212}{−}
  \usepgfplotslibrary{groupplots,dateplot}
  \usetikzlibrary{patterns,shapes.arrows}
  \usetikzlibrary{fit,shapes,calc,positioning}
  \pgfplotsset{compat=newest}

  \usepackage{footnote}
  
\def\mat#1{\boldsymbol{#1}}
\def\vec#1{\boldsymbol{#1}}
\def\set#1{\{#1\}}
\def\tuple#1{(#1)}
\def\lest#1{\tuple{#1}}

\def\pct#1{\qty{#1}{\percent}}

\newcommand{\legendbox}[2]{{
{
\tcbox[
  colframe=black,
  colback=#1,
  size=fbox,
  arc=1pt,
  outer arc=1.4pt,
  boxrule=0.4pt,
  nobeforeafter,
  boxsep=0pt
]{\rule{1.0em}{0em}\rule{0em}{1.20ex}}%
}{\,#2}}}

\newcommand{\diagbox}[3]{{%
{
\tcbox[
  enhanced,
  colframe=black,
  size=fbox,
  nobeforeafter,
  boxrule=0.4pt,
  arc=1pt,
  outer arc=1.4pt,
  boxsep=0pt,
  underlay={%
    \begin{tcbclipinterior}
    \fill[#1] (interior.south west) -- (interior.north west) -- (interior.north east) -- cycle;
    \fill[#2] (interior.south west) -- (interior.south east) -- (interior.north east) -- cycle;
    \end{tcbclipinterior}
  }
  ]{\rule{1.0em}{0em}\rule{0em}{1.20ex}}%
}}{\,#3}}

\DeclareRobustCommand{\colorcircle}[1]{\tikz[baseline=-0.5ex]{\draw[fill=#1,draw=black,line width=0.3pt] (0,0) circle (0.22em);}}
\DeclareRobustCommand{\colorline}[1]{\tikz[baseline=-0.5ex]{\draw[draw=#1,line width=1.5pt] (0,0) -- (0.95em,0);}}
\DeclareRobustCommand{\colorlinecircle}[1]{\tikz[baseline=-0.5ex]{\draw[draw=#1,line width=1.5pt] (0,0) -- (0.95em,0);\draw[fill=#1,draw=black,line width=0.3pt] (1.0em,0) circle (0.20em);}}

\DeclareRobustCommand{\eat}[1]{}

\def\tymes{{\times}}

\def\figcontents#1{\centering#1\vskip -2.5mm}

\def\ds{KTH floor plan dataset}
\def\eg{e.g.}
\def\ie{i.e.}

\def\GT#1{{\hat{#1}}}
\def\Est#1{\tilde{#1}}
\def\MLE#1{{#1^*}}
\def\P#1{p(#1)}
\def\PC#1#2{\P{#1 \mid #2}}
\DeclareRobustCommand{\[}{\begin{equation}}
\DeclareRobustCommand{\]}{\end{equation}}
\def\eqnend#1{}
\def\VertexSet{\ensuremath{\mathcal{V}}}
\def\TokenSet{\ensuremath{\mathcal{T}}}
\def\CellStates{\ensuremath{\mathcal{C}}}
\def\ax{a}\def\bx{\alpha}
\def\ay{b}\def\by{\beta}
\def\ax{x}\def\bx{x'}
\def\ay{y}\def\by{y'}
\def\qv{q}

\def\R{\mathbb{R}}
\def\constant#1{\ifmmode\text{`\MakeLowercase{#1}'}\else`\MakeLowercase{#1}'\fi}
\def\SOS{\constant{Start}}
\def\EOS{\constant{End}}
\def\U{\constant{Unknown}}
\def\F{\constant{Free}}
\def\O{\constant{Occupied}}
\def\W{\constant{Window}}
\def\UFOW{\U{},\F{},\O{},\W}
\def\Iverson#1{\left[#1\right]}
\def\Expect#1{\left\langle#1\right\rangle}
\def\floor#1{\lfloor{#1}\rfloor}

\DeclareSIUnit\cells{cells}

\definecolor{changecolor}{RGB}{50,0,225} 
\newenvironment{changed}{}{}
\def\change#1{{#1}}

  \definecolor{mplC0}{HTML}{1F77B4}
\definecolor{mplC1}{HTML}{FF7F0E}
\definecolor{mplC2}{HTML}{2CA02C}
\definecolor{mplC3}{HTML}{D62728}
\definecolor{mplC4}{HTML}{9467BD}
\definecolor{mplC5}{HTML}{8C564B}
\definecolor{mplC6}{HTML}{E377C2}
\definecolor{mplC7}{HTML}{7F7F7F}
\definecolor{mplC8}{HTML}{BCBD22}
\definecolor{mplC9}{HTML}{17BECF}
\definecolor{mplaliceblue}{HTML}{F0F8FF}
\definecolor{mplantiquewhite}{HTML}{FAEBD7}
\definecolor{mplaqua}{HTML}{00FFFF}
\definecolor{mplaquamarine}{HTML}{7FFFD4}
\definecolor{mplazure}{HTML}{F0FFFF}
\definecolor{mplbeige}{HTML}{F5F5DC}
\definecolor{mplbisque}{HTML}{FFE4C4}
\definecolor{mplblack}{HTML}{000000}
\definecolor{mplblanchedalmond}{HTML}{FFEBCD}
\definecolor{mplblue}{HTML}{0000FF}
\definecolor{mplblueviolet}{HTML}{8A2BE2}
\definecolor{mplbrown}{HTML}{A52A2A}
\definecolor{mplburlywood}{HTML}{DEB887}
\definecolor{mplcadetblue}{HTML}{5F9EA0}
\definecolor{mplchartreuse}{HTML}{7FFF00}
\definecolor{mplchocolate}{HTML}{D2691E}
\definecolor{mplcoral}{HTML}{FF7F50}
\definecolor{mplcornflowerblue}{HTML}{6495ED}
\definecolor{mplcornsilk}{HTML}{FFF8DC}
\definecolor{mplcrimson}{HTML}{DC143C}
\definecolor{mplcyan}{HTML}{00FFFF}
\definecolor{mpldarkblue}{HTML}{00008B}
\definecolor{mpldarkcyan}{HTML}{008B8B}
\definecolor{mpldarkgoldenrod}{HTML}{B8860B}
\definecolor{mpldarkgray}{HTML}{A9A9A9}
\definecolor{mpldarkgreen}{HTML}{006400}
\definecolor{mpldarkgrey}{HTML}{A9A9A9}
\definecolor{mpldarkkhaki}{HTML}{BDB76B}
\definecolor{mpldarkmagenta}{HTML}{8B008B}
\definecolor{mpldarkolivegreen}{HTML}{556B2F}
\definecolor{mpldarkorange}{HTML}{FF8C00}
\definecolor{mpldarkorchid}{HTML}{9932CC}
\definecolor{mpldarkred}{HTML}{8B0000}
\definecolor{mpldarksalmon}{HTML}{E9967A}
\definecolor{mpldarkseagreen}{HTML}{8FBC8F}
\definecolor{mpldarkslateblue}{HTML}{483D8B}
\definecolor{mpldarkslategray}{HTML}{2F4F4F}
\definecolor{mpldarkslategrey}{HTML}{2F4F4F}
\definecolor{mpldarkturquoise}{HTML}{00CED1}
\definecolor{mpldarkviolet}{HTML}{9400D3}
\definecolor{mpldeeppink}{HTML}{FF1493}
\definecolor{mpldeepskyblue}{HTML}{00BFFF}
\definecolor{mpldimgray}{HTML}{696969}
\definecolor{mpldimgrey}{HTML}{696969}
\definecolor{mpldodgerblue}{HTML}{1E90FF}
\definecolor{mplfirebrick}{HTML}{B22222}
\definecolor{mplfloralwhite}{HTML}{FFFAF0}
\definecolor{mplforestgreen}{HTML}{228B22}
\definecolor{mplfuchsia}{HTML}{FF00FF}
\definecolor{mplgainsboro}{HTML}{DCDCDC}
\definecolor{mplghostwhite}{HTML}{F8F8FF}
\definecolor{mplgold}{HTML}{FFD700}
\definecolor{mplgoldenrod}{HTML}{DAA520}
\definecolor{mplgray}{HTML}{808080}
\definecolor{mplgreen}{HTML}{008000}
\definecolor{mplgreenyellow}{HTML}{ADFF2F}
\definecolor{mplgrey}{HTML}{808080}
\definecolor{mplhoneydew}{HTML}{F0FFF0}
\definecolor{mplhotpink}{HTML}{FF69B4}
\definecolor{mplindianred}{HTML}{CD5C5C}
\definecolor{mplindigo}{HTML}{4B0082}
\definecolor{mplivory}{HTML}{FFFFF0}
\definecolor{mplkhaki}{HTML}{F0E68C}
\definecolor{mpllavender}{HTML}{E6E6FA}
\definecolor{mpllavenderblush}{HTML}{FFF0F5}
\definecolor{mpllawngreen}{HTML}{7CFC00}
\definecolor{mpllemonchiffon}{HTML}{FFFACD}
\definecolor{mpllightblue}{HTML}{ADD8E6}
\definecolor{mpllightcoral}{HTML}{F08080}
\definecolor{mpllightcyan}{HTML}{E0FFFF}
\definecolor{mpllightgoldenrodyellow}{HTML}{FAFAD2}
\definecolor{mpllightgray}{HTML}{D3D3D3}
\definecolor{mpllightgreen}{HTML}{90EE90}
\definecolor{mpllightgrey}{HTML}{D3D3D3}
\definecolor{mpllightpink}{HTML}{FFB6C1}
\definecolor{mpllightsalmon}{HTML}{FFA07A}
\definecolor{mpllightseagreen}{HTML}{20B2AA}
\definecolor{mpllightskyblue}{HTML}{87CEFA}
\definecolor{mpllightslategray}{HTML}{778899}
\definecolor{mpllightslategrey}{HTML}{778899}
\definecolor{mpllightsteelblue}{HTML}{B0C4DE}
\definecolor{mpllightyellow}{HTML}{FFFFE0}
\definecolor{mpllime}{HTML}{00FF00}
\definecolor{mpllimegreen}{HTML}{32CD32}
\definecolor{mpllinen}{HTML}{FAF0E6}
\definecolor{mplmagenta}{HTML}{FF00FF}
\definecolor{mplmaroon}{HTML}{800000}
\definecolor{mplmediumaquamarine}{HTML}{66CDAA}
\definecolor{mplmediumblue}{HTML}{0000CD}
\definecolor{mplmediumorchid}{HTML}{BA55D3}
\definecolor{mplmediumpurple}{HTML}{9370DB}
\definecolor{mplmediumseagreen}{HTML}{3CB371}
\definecolor{mplmediumslateblue}{HTML}{7B68EE}
\definecolor{mplmediumspringgreen}{HTML}{00FA9A}
\definecolor{mplmediumturquoise}{HTML}{48D1CC}
\definecolor{mplmediumvioletred}{HTML}{C71585}
\definecolor{mplmidnightblue}{HTML}{191970}
\definecolor{mplmintcream}{HTML}{F5FFFA}
\definecolor{mplmistyrose}{HTML}{FFE4E1}
\definecolor{mplmoccasin}{HTML}{FFE4B5}
\definecolor{mplnavajowhite}{HTML}{FFDEAD}
\definecolor{mplnavy}{HTML}{000080}
\definecolor{mploldlace}{HTML}{FDF5E6}
\definecolor{mplolive}{HTML}{808000}
\definecolor{mplolivedrab}{HTML}{6B8E23}
\definecolor{mplorange}{HTML}{FFA500}
\definecolor{mplorangered}{HTML}{FF4500}
\definecolor{mplorchid}{HTML}{DA70D6}
\definecolor{mplpalegoldenrod}{HTML}{EEE8AA}
\definecolor{mplpalegreen}{HTML}{98FB98}
\definecolor{mplpaleturquoise}{HTML}{AFEEEE}
\definecolor{mplpalevioletred}{HTML}{DB7093}
\definecolor{mplpapayawhip}{HTML}{FFEFD5}
\definecolor{mplpeachpuff}{HTML}{FFDAB9}
\definecolor{mplperu}{HTML}{CD853F}
\definecolor{mplpink}{HTML}{FFC0CB}
\definecolor{mplplum}{HTML}{DDA0DD}
\definecolor{mplpowderblue}{HTML}{B0E0E6}
\definecolor{mplpurple}{HTML}{800080}
\definecolor{mplrebeccapurple}{HTML}{663399}
\definecolor{mplred}{HTML}{FF0000}
\definecolor{mplrosybrown}{HTML}{BC8F8F}
\definecolor{mplroyalblue}{HTML}{4169E1}
\definecolor{mplsaddlebrown}{HTML}{8B4513}
\definecolor{mplsalmon}{HTML}{FA8072}
\definecolor{mplsandybrown}{HTML}{F4A460}
\definecolor{mplseagreen}{HTML}{2E8B57}
\definecolor{mplseashell}{HTML}{FFF5EE}
\definecolor{mplsienna}{HTML}{A0522D}
\definecolor{mplsilver}{HTML}{C0C0C0}
\definecolor{mplskyblue}{HTML}{87CEEB}
\definecolor{mplslateblue}{HTML}{6A5ACD}
\definecolor{mplslategray}{HTML}{708090}
\definecolor{mplslategrey}{HTML}{708090}
\definecolor{mplsnow}{HTML}{FFFAFA}
\definecolor{mplspringgreen}{HTML}{00FF7F}
\definecolor{mplsteelblue}{HTML}{4682B4}
\definecolor{mpltan}{HTML}{D2B48C}
\definecolor{mplteal}{HTML}{008080}
\definecolor{mplthistle}{HTML}{D8BFD8}
\definecolor{mpltomato}{HTML}{FF6347}
\definecolor{mplturquoise}{HTML}{40E0D0}
\definecolor{mplviolet}{HTML}{EE82EE}
\definecolor{mplwheat}{HTML}{F5DEB3}
\definecolor{mplwhite}{HTML}{FFFFFF}
\definecolor{mplwhitesmoke}{HTML}{F5F5F5}
\definecolor{mplyellow}{HTML}{FFFF00}
\definecolor{mplyellowgreen}{HTML}{9ACD32}

\begin{document}


  \def\t{Beyond the Frontier: Predicting Unseen Walls from Occupancy Grids by Learning from Floor Plans}
  \def\a{Ludvig Ericson, Patric Jensfelt}

  \title{\t}


  \ifhyperref
    \hypersetup{%
      pdftitle=\t,
      pdfauthor=\a,
      pdfcreator=,
      bookmarksopenlevel=2
    }
  \fi

  \newif\ifabstract{}
  \newif\iffigontop{}
  \newif\iffinal{}

  \abstracttrue{}
  \figontoptrue{}
  \finalfalse{}

  \author{\a{}%
    \thanks{This work was supported by the Swedish Research Council.
    All authors are with the Division of Robotics, Perception and Learning at KTH Royal Institute of Technology, Stockholm, SE-10044, Sweden. For e-mail correspondence, contact {\tt\footnotesize{ludv@kth.se}}.}%
  }

  \iffinal%
    \markboth%
    {IEEE Robotics and Automation Letters. Preprint Version. Accepted May, 2024}%
    {L. Ericson, P. Jensfelt: \t{}} 
  \else
    \pagenumbering{gobble}
  \fi

  \maketitle


  \ifabstract{}
  \begin{abstract}

  \begin{changed}
    In this paper, we tackle the challenge of predicting the unseen walls of a
    partially observed environment as a set of 2D line segments, conditioned on
    occupancy grids integrated along the trajectory of a \ang{360} LIDAR
    sensor. A dataset of such occupancy grids and their corresponding target
    wall segments is collected by navigating a virtual robot between a set of
    randomly sampled waypoints in a collection of office-scale floor plans from
    a university campus.
    The line segment prediction task is formulated as an autoregressive
    sequence prediction task, and an attention-based deep network is trained on
    the dataset. The sequence-based autoregressive formulation is evaluated
    through predicted information gain, as in frontier-based autonomous
    exploration, demonstrating significant improvements over both
    non-predictive estimation and convolution-based image prediction found in
    the literature. Ablations on key components are evaluated, as well as
    sensor range and the occupancy grid's metric area. Finally, model
    generality is validated by predicting walls in a novel floor plan
    reconstructed on-the-fly in a real-world office environment.
  \end{changed}

  \end{abstract}
  \fi


  \def\igrefs{zhou2021fuel,shrestha2019learned}
  \def\baselinerefs{shrestha2019learned,katyal2019uncertainty,zwecher2022integrating,tao2023learning,tao2023seer}
  \def\citerhnbv{\cite{bircher2016receding}}%
  \def\citeunet{\cite{tao2023seer,katyal2019uncertainty}}%

\section{Introduction}

  \IEEEPARstart{H}{uman} and robotic problem-solving approaches differ in
  dealing with the unknown and predicting the near future. Classical robotic
  approaches seek exactness at the cost of intuition and foresight, and on the
  contrary, humans do not meticulously maintain metric maps of their worlds.
  Even schematics and blueprints, documents explicitly intended to specify
  technical details, leave room for interpretation. Abstraction seems necessary
  for our ability to move between the specifics of reality and the generality
  of ideas and ideals. Replicating this ability to reason abstractly with
  explicit algorithms has historically proven difficult, but recent advances in
  learning-based approaches have opened up new avenues and great strides have
  been made across many subfields of robotics.

  In this work, floor plans are used as the medium through which abstract
  reasoning is made possible. Floor plans are the architectural blueprints of
  our built environment, a distillate of the real world as a tidy set of shapes
  and symbols encoding the layouts and purposes of rooms, positions of walls,
  doorways, and windows. Floor plans obey rules, symmetries, and regularities
  that are impossible to state explicitly, \change{often driven by aesthetic
  considerations rather than logic. We have previously shown that recent
  advances in autoregressive language models can be leveraged to produce a
  generative model over floor plans as sequences of vector graphic
  instructions~\cite{iros22}. By contrast to single-shot approaches such as
  \citeunet{}, an autoregressive approach casts the floor plan generation task
  as a series of decisions and their consequences, enabling the model to reason
  in steps, analogous to the chain-of-thought paradigm in large language
  models~\cite{wei2022chain}.}

  \change{Autonomous exploration planning is an obvious example of a classical
  robotics problem where such a prediction model should be immediately
  applicable; however, we have previously shown that traditional non-predictive
  exploration planners are not well-suited to using predictions, and
  predictions can actually have a negative impact on exploration
  performance~\cite{ecmr21}. In this article, we have limited the scope to
  evaluating predictions by the primary variable that they affect in the
  exploration context: \emph{predicted information gain}.}

  This paper is structured as follows.
  In \cref{sec:probmodel}, an model for predicting floor plans from sensor
  history, dubbed \emph{Floorist}, is defined. Data modality is the main
  difference from our previous model~\cite{iros22}, which dealt solely with
  abstract floor plans. Floorist is instead grounded in the real world by
  taking a partially observed environment as input in the form of 2D occupancy
  grids from a \ang{360} LIDAR sensor and \change{predicting the unobserved
  walls of that environment as line segments.
  In \cref{sec:datagen}, a dataset generation method is outlined wherein a
  virtual robot navigates between randomly sampled waypoints in a collection of
  annotated floor plans, generating input occupancy grids and their target wall
  segments.
  In \cref{sec:pig}, cluster-based predicted information gain is defined as
  in~\cite\igrefs{}, it is the evaluation metric used in this work, suitable
  for occupancy grid-based prediction models.
  In \cref{sec:expsetup,sec:expresults}, three prediction models are evaluated:
  Floorist, a baseline convolution-based architecture as
  in~\cite\baselinerefs{}, and a non-predictive approach as in~\citerhnbv{}.
  Floorist performance under ablation of key components is also reported, as
  well as its sensitivity to sensor range and occupancy grid area.
  Finally, in \cref{sec:realworld}, Floorist is applied to an on-the-fly floor
  plan reconstruction of a real-world office environment to validate its
  generality.}

  \def\predsegs{\protect\colorline{mplC1}}
  \def\truesegs{\protect\colorline{mplC0}}
  \def\trajline{\protect\colorlinecircle{mplC6}}

  \newif\iftimestamp
  \timestampfalse

  \begin{figure}[bt]
    \iftimestamp%
      \makebox[\linewidth]{%
        \footnotesize%
        \def\frame#1{{#1}}%
        \def\w{0.333\linewidth}%
        \hspace*{\fill}%
        \frame{\makebox[\w]{(a) $t=0$}}%
        \frame{\makebox[\w]{(b) $t=1$}}%
        \frame{\makebox[\w]{(c) $t=2$}}%
        \hspace*{\fill}%
      } \\%
    \fi%
    \begin{adjustbox}{clip,trim=2mm,width=\linewidth}%
      \input{murder_in_three_acts/murder_in_three_acts.pgf}%
    \end{adjustbox}%
    \caption{Three consecutive occupancy grids with
    \legendbox{mplgray}{Unknown},
    \legendbox{white}{Free},
    \legendbox{mplblue}{Occu\-pied}, and
    \legendbox{mplturquoise}{Window} cells;
    \predsegs{}~Predicted walls from a Floorist model;
    \truesegs{}~Target walls; and 
    \trajline{}~Trajectory. Initially (left), few lines match up
    exactly, apart from the northern exterior wall which is visible, and the
    predicted rooms do exist though not in the exact locations predicted. In
    the next step (middle), more information about the corridor is observed,
    and the predicted segments are also improved, though the adjoining rooms
    are still misaligned and their doorways misplaced. Finally (right), the
    adjacent rooms are partially observed, and the predicted doorway alignment
    is now correct, and both room's widths are correctly adjusted. The images
    have been cropped for legibility.}%
    \label{fig:murder}%
  \end{figure}

  \def\mylink{\href{https://lericson.se/floorist/}{\nolinkurl{lericson.se/floorist}}}
  \def\spicylink#1{\mbox{{\ttfamily{}\bfseries{}\footnotesize{}#1}}}

  The dataset generation software along with training and inference code for
  Floorist is published in tandem with this work under an open-source license
  at \spicylink\mylink{}.


\section{Related Work}

  \def\aerefs{zhou2021fuel,duberg2022ufoexplorer,yu2023echo}
  \def\ogmrefs{tao2023seer,shrestha2019learned,katyal2019uncertainty,zwecher2022integrating,tao2023learning}
  \def\samplingrefs{bircher2016receding,selin2019efficient,schmid2022fast}

  The approach of predicting the unknown from occupancy grids is perhaps most
  common in the autonomous exploration literature. Autonomous exploration is
  the task of reconstructing a map of an environment, typically with no prior
  information about that particular environment, \eg{}, \cite\aerefs{}, though
  not always~\cite{luperto2020robot}. In \emph{frontier-based} exploration
  planners, the planner considers \emph{frontiers} by spatial clustering of the
  boundary between free and unknown space. A score is assigned to each
  frontier, and the planner navigates to the highest-scoring frontier. The
  score is typically a function of the travel distance and the predicted
  information gain, an estimate of how many bits of new information will be
  obtained by visiting that frontier.
  A popular predictive approach, \change{\eg{}, \cite\ogmrefs{}}, is to derive
  training data directly from autonomous exploration planning and then train
  some neural network to predict a 2D occupancy grid, computing the predicted
  information gain from the occupancy grid prediction.
  \change{In sampling-based exploration planners such as \cite\samplingrefs{},
  path and exploration planning are performed simultaneously with some variant
  of RRT~\cite{karaman2011sampling}.} Some sampling-based works predict
  information gain directly by a regression formulation, with a deep
  network~\cite{schmid2022fast} or a Gaussian
  process~\cite{selin2019efficient}. \change{In \cite{zwecher2022integrating},
  a reinforcement learning-based approach is proposed paired with a
  convolutional network for occupancy grid prediction; \cite{tao2023learning}
  extends the approach to a real-world exploration system in the form of a
  micro-aerial vehicle.}

  Another line of inquiry is the explicit reconstruction of floor plans from
  sensor data for architectural purposes, typically in an offline setting.
  In~\cite{yue2023roomformer,su2023slibo,chen2022heat} use an attention-based
  network to approach the task of predicting a cohesive floor plan given a set
  of point clouds covering the entire environment. \cite{gueze2023floor} solve
  a similar task, though using sparse multi-views instead of point clouds.
  Their approach is to model the environment topologically before generating a
  floor plan suitable to those topological constraints.
  In~\cite{luperto2023mapping}, the aim is somewhere between offline and online
  prediction, posing the problem of inferring what is behind closed doors given
  an occupancy grid of the observable parts of the environment, \ie{}, after
  exploration is completed. The authors demonstrate favorable performance in
  path planning tasks into unknown space, behind closed doors.

  Some works concerning autonomous driving take a similar approach to ours in
  representing and predicting the environment as a set of line segments with a
  neural network~\cite{liao2023maptr,li2021hdmapnet}. Emphasis is placed on
  reconstruction of the visible environment as opposed to reasoning about the
  unknown, though many of the technical challenges are similar. In the category
  of learning-based approaches on vector graphics, the approach is typically to
  embed an entire graphic to enable downstream tasks on that embedding, such as
  animation and interpolation~\cite{carlier2020deepsvg,reddy2021im2vec}.
  
  In~\cite{aydemir}, a topological approach to modeling and reasoning about
  indoor environnments is taken. It also introduces the \emph{\ds{}}, which is
  used for synthesizing training data in the present work. It is a collection
  of floor plans from a university campus, annotated with positions of walls,
  doors, and windows.

  
\section{Notation}


  $\Expect{\cdot}$ denotes expectation under some probability distribution.
  $\Iverson{\cdot}$ denotes the Iverson bracket, sometimes called the indicator
  function. It maps the truth value of a proposition to one or zero.
  $\Expect{\Iverson{x=1}}$ thus denotes the probability that $x = 1$.
  $\floor{x}$ denotes the floor function, \ie{}, the greatest integer less than
  or equal to $x$.

\section{Probabilistic Model}\label{sec:probmodel}

    \def\n{\mathrm{N}}
    \def\LineSegMats{\R^{N\tymes{}4}}

  The problem of predicting floor plans is formulated as an autoregressive
  prediction task on sequences of line segment vertices. The perspective is an
  in-situ robot located at the origin having observed some part of its
  surroundings, and the task is to predict the walls of the floor plan beyond
  what has already been observed, as illustrated in \cref{fig:murder}.
    The floor plan is represented as a matrix $\mat{S} \in \LineSegMats$ of
    the $N$ target line segments from $\tuple{\ax{},\ay{}}$ to
    $\tuple{\bx{},\by{}}$, \ie, \[
    \mat{S} = \begin{pmatrix} \ax_1 & \ay_1 & \bx_1 & \by_1 \\ 
                              \ax_2 & \ay_2 & \bx_2 & \by_2 \\ 
                              \vdots & \vdots & \vdots & \vdots \\
                              \ax_\n & \ay_\n & \bx_\n & \by_\n \end{pmatrix}
    \] The vector graphic produced by $\mat{S}$ is invariant to row-wise
    permutations, and vertex order reversal. However, in an autoregressive
    regime, order does matter and must be chosen carefully for two reasons.
    First, any function that models an ordered sequence must by necessity
    subsume the ordering algorithm. Secondly, autoregression implies that later
    positions in the sequence are informed by earlier positions. We use a
    proximity heuristic where the rows are ordered by the distance from the
    robot to the nearest point on \change{each line segment, as in
    \cite{iros22}. The intuition is that segments near the robot are often
    partially observed, and can be predicted using the occupancy grid as
    grounding. Later segments that are far away and ungrounded can then be
    chosen to fit with the nearby grounded segments.}
    For vertex order, a simple strategy suffices. The vertices of each segment
    are ordered lexicographically, \ie,
    \[ (\ax{} < \bx{}) \lor ((\ax{} = \bx{}) \land (\ay{} \le \by{})) \]

  \def\vecivsub#1_#2{\vec{#1}_{\Iverson{#2}}}
  \def\tS{\vec{t}(\mat{S})}
  \def\tSM{\vec{t}(\mat{S}(\mat{M}))}
  \subsection{Sequence Tokenization and Factorization}
    The line segment prediction problem is cast as a sequence prediction
    problem on the \emph{token sequence} $\vec{t}$ where the joint distribution
    is conditioned on some contextual input $C$, that is \[
      \PC{\mat{S}}{C} = \PC{\vec{t}}{C} \label{eq:recur}
    \] The sequence $\vec{t}$ is initialized and terminated by a \SOS{} and
    \EOS{} token respectively, and otherwise consists of vertex pairs, each
    vertex is quantized \change{by $q : \R\tymes\R \to \mathcal{V}$ with the index set
    $\VertexSet=\set{1, 2, \ldots, HW}$ and} \[
    \label{eq:quantize}
      \qv(x, y) = \floor{ W(\tfrac{1}{2}H - s_y y) + (\tfrac{1}{2}W + s_x x)}
    \] for some $H\tymes{}W$ grid at scale $s_x, s_y$ in
    \unit{\cells\per\meter}. The function from segments $\mat{S}$ to
    \change{tokens $\tS{} \in \TokenSet^{2N+2}$ with the token vocabulary
    $\TokenSet = \set{\SOS{}, \EOS{}} \cup \VertexSet$ is then defined} \[
    \label{eq:sequencing} \begin{aligned}
      \tS{} = \lest{\SOS{}, {}& \qv(\ax_1, \ay_1), \qv(\bx_1, \by_1),           \\ 
                                       {}& \qv(\ax_2, \ay_2), \qv(\bx_2, \by_2), \, \ldots \\
                                       {}& \qv(\ax_\n, \ay_\n), \qv(\bx_\n, \by_\n), \EOS{}}
    \end{aligned} \]
    The joint probability is factorized autoregressively as in the
    recursion \[ \label{eq:factorization} \begin{aligned}
      \PC{\vecivsub{t}_{j\le{}i}}{C} &= \PC{t_i}{\vecivsub{t}_{j<i}, C} \, \PC{\vecivsub{t}_{j<i}}{C} \\
      \PC{\vecivsub{t}_{j\le{}1}}{C} &= \Iverson{t_1 = \SOS{}}
    \end{aligned} \] where $t_i$ denotes the $i$th token, $\vecivsub{t}_{j<i}$
    and $\vecivsub{t}_{j\le{}i}$ denote the subsequence of $\vec{t}$ up to but
    excluding (or including) $i$.
    The next-token distribution is parameterized by logits from a deep network
    $\vec{f}_\theta$, \ie, \[ \label{eq:network}
      \PC{t_i}{\vecivsub{t}_{j<i}, C} = \sigma(\vec{f}_\theta(\vecivsub{t}_{j<i}, C) )_i
    \] where $\sigma(\cdot)_i$ is the $i$th element of the normalized logistic
    function. The function $\vec{f}_\theta$ assigns probability mass to the
    true token sequence $\GT{\vec{t}}$ by gradient descent on $\Expect{L}$, the expected
    negative log likelihood under the data distribution, with \[ \label{eq:loss}
      L_\theta(\GT{\vec{t}}, C) = -\sum_{\GT{t}_i\in\GT{\vec{t}}} \log \sigma(\vec{f}_\theta(\vecivsub{\GT{t}}_{j<i}, C))_{\GT{t}_i}
    \]

  \subsection{Contextual Inputs}

    \def\OGs{\CellStates^{H\tymes{}W}}

    The contextual input $C$ consists of two parts: a partial occupancy grid
    $\mat{M} \in \OGs$, and the \change{\emph{visible line segments}
    $\mat{S}(\mat{M}) \in \LineSegMats$} recovered from the occupancy grid. The
    cell labels are \[ \CellStates{}=\set{\UFOW{}} \label{eq:cellstates} \]
    \change{ The label \W{} indicates that a cell contains an \emph{exterior window},
    \ie, a window facing outside the building, and hints the floor plan's
    perimeter.
    Marching squares~\cite{maple2003ms} is used to find the visible line
    segments $\mat{S}(\mat{M})$.
    In regions where at least one cell is marked \U{}, no line segment is
    produced.
    }

  \subsection{Network Architecture}

    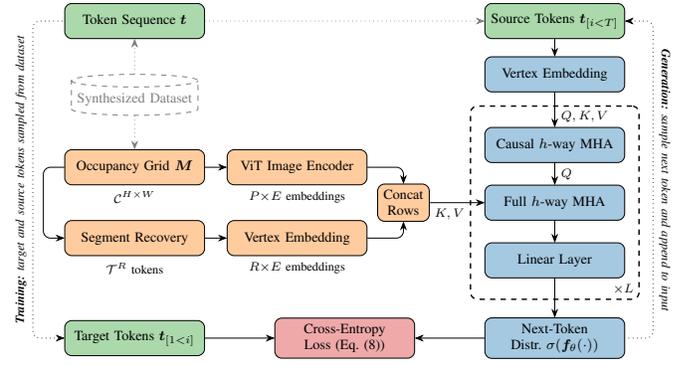
\begin{figure}
      \vspace{1mm}
      \tikzset{
        node distance=2em and 8mm,
        every node/.style={font=\footnotesize},
        label style/.style={font=\scriptsize},
        basic/.style={text width=7.3em, text centered},
        block/.style={basic, rectangle, draw, fill=none, rounded corners, minimum height=2em},
        line/.style={draw, rounded corners, ->},
        cloud/.style={draw, ellipse,fill=red!20, minimum height=2em},
        encoder/.style={block, fill=mplC1!40},
        input/.style={block, fill=mplC2!40},
        layer/.style={block, fill=mplC0!40},
      }

      \begin{adjustbox}{max width=\linewidth}
        \begin{tikzpicture}[>=Stealth]


        \node[encoder, label={[label style]below:$\CellStates^{H\tymes{}W}$}] (ogm) {Occupancy Grid $\mat{M}$};
        \node[encoder, right=1.25em of ogm, label={[label style]below:$P\tymes{}E$ embeddings}] (vit) {ViT Image Encoder};
        \node[encoder, below=of ogm, label={[label style]below:$\TokenSet{}^{R}$ tokens}] (linesegs) {Segment Recovery};
        \node[encoder, right=1.25em of linesegs, label={[label style]below:${R}\tymes{E}$ embeddings}] (vertemb) {Vertex Embedding};
        \node[encoder, right=2mm of $(vit.east)!0.5!(vertemb.east)$, text width=8mm] (concat) {Concat Rows};
        \node[layer, right=11mm of concat] (crossatt) {Full $h$-way MHA};
        \node[layer, above=1.25em of crossatt] (selfatt) {Causal $h$-way MHA};
        \node[layer, below=1.25em of crossatt] (mlp) {Linear Layer};

        \node[draw=black, dashed, fit=(crossatt) (selfatt) (mlp), inner xsep=8pt, inner ysep=1.2em, thick, rectangle, rounded corners, label={[label style,anchor=south east]south east:$\tymes{}L$}] (group) {};

        \node[layer, below=1em of group] (output) {Next-Token Distr. $\sigma(\vec{f}_\theta(\cdot))$};
        \node[layer, above=0.75em of group] (vertemb2) {Vertex Embedding};
        \node[input, above=1em of vertemb2] (input) {Source Tokens $\vec{t}_{\Iverson{i<T}}$};

        \path (ogm |- input) node[input] (refseq) {Token Sequence $\vec{t}$};
        \node[cylinder, below=of refseq,draw,dashed,gray,aspect=0.1,fill=none,minimum height=2em,minimum width=7em,shape border rotate=90] (floorplan) {Synthesized Dataset};
        \path[line, dotted, gray] (floorplan) -- (ogm);
        \path[line, dotted, gray] (floorplan) -- (refseq);
        \path[line, dotted, gray] (refseq) -- (input);

        \path[line] (ogm) -- (vit);
        \path[line] (ogm.west) -- ([xshift=-12pt]ogm.west) -- ([xshift=-12pt]linesegs.west) -- (linesegs.west);
        \path[line] (linesegs) -- (vertemb);
        \path[line] (vit) -| (concat);
        \path[line] (vertemb) -| (concat);
        \path[line] (concat) -- (crossatt);
        \path (concat.east) -- (crossatt.west -| group.west) node[midway, below] {\scriptsize $K,V$};
        \path[line] (crossatt) -- (mlp);
        \path[line] (mlp) -- (output);
        \path[line] (input) -- (vertemb2);
        \path[line] (vertemb2) -- (selfatt);
        \path (group.north) -- (selfatt.north) node[midway, right] {\scriptsize $Q, K, V$};
        \path[line] (selfatt) -- (crossatt) node[midway, right] {\scriptsize $Q$};

        \path[line, dotted, <-] (input.east) -- ([xshift=16pt]input.east) -- ([xshift=16pt]output.east) node[midway, sloped, above] {\scriptsize\itshape {\bfseries{}Generation:} sample next token and append to input} -- (output.east);

        \path (ogm |- output) node[input] (target) {Target Tokens $\vec{t}_{\Iverson{1<i}}$}; 
        \path (target) -- (output) node[midway, block, fill=mplC3!40] (loss) {Cross-Entropy Loss (\cref{eq:loss})};
        \path[line] (target) -- (loss);
        \path[line] (output) -- (loss);

        \path[line, dotted, <-] (target.west) -- ([xshift=-18pt]target.west) -- ([xshift=-18pt]refseq.west) node[sloped, midway, above] {\scriptsize\itshape {\bfseries{}Training:} target and source tokens sampled from dataset} -- (refseq.west);

        \end{tikzpicture}%
      \end{adjustbox}%
      \caption{The Floorist network architecture with training and generation
      pathways. In training, source and target token sequences are samples from
      the dataset, where \legendbox{mplC1!40}{Encoder blocks} and
      \legendbox{mplC0!40}{Decoder blocks} are evaluated in one pass. In
      generation, the source sequence starts as $\SOS{}$ and the encoder side
      is only evaluated once. The decoder is then evaluated to obtain the
      next-token distribution, and a token is sampled from it and appended to
      the source sequence before the process repeats again.
      $R=|\tSM{}|$ is the number of tokens in the line segments
      visible in $\mat{M}$, $T=|\vec{\GT{t}}|$ is the number of target tokens.
      $P$ is the number of patches from ViT, $E$ is the embedding dimension,
      $Q,K,V$ are the query, key, and value matrices for a multi-head attention
      (MHA) layer. $\vec{t}_{\Iverson{i<T}}$ denotes removal of the last token
      (\ie, \EOS{}), and $\vec{t}_{\Iverson{1<i}}$ denotes removal of the first
      token (\ie, \SOS{}). This shifts the two sequences so that the target is
      always the next token. Note that each attention block is a
      gated residual connection as in~\cite{rezero}.}\label{fig:arch}
    \end{figure}

    \change{Following success in the language modeling domain, the choice of
    $f_\theta$ in \cref{eq:network}} is a transformer encoder-decoder with
    multi-head attention (MHA) as in~\cite{aiayn}, illustrated in
    \cref{fig:arch}. The encoder is computed once per sequence to encode the
    contextual input, while the decoder is iteratively re-evaluated
    step-by-step during sampling as the generated sequence is constructed.
    Unlike~\cite{aiayn}, each transformer layer is a residual layer with a
    gated residual as in~\cite{rezero}.
    The two modalities of the contextual input are tokenized separately. The
    occupancy grid $\mat{M}$ is encoded using a ViT
    network~\cite{dosovitskiy2020vit}, and the tokenized visible line segments
    $\tSM$ are encoded with a discrete embedding with absolute position
    encoding. The context tokens are concatenated along the sequence dimension
    and cross-attended to.

    \def\bits#1{\texttt{#1}}
    \def\cvec#1#2#3{{\left[ {#1}1; {#2}1; {#3}1 \right]}}
    \def\Chroma{k}

    \begin{changed}
    \subsubsection{Chromatization}\label{sec:chroma} Cell labels
    are mapped to 3-channel pseudo-colors by $\Chroma{} : \CellStates{} \to \R{}^3$ with \[
      \nonumber \begin{aligned}
        \Chroma{}(\U{}) &= \cvec--- &
        \Chroma{}(\F{}) &= \cvec-+- \\
        \Chroma{}(\O{}) &= \cvec--+ &
        \Chroma{}(\W{}) &= \cvec+-+
      \end{aligned}
    \] %
    The exact choice of pseudo-colors is inconsequential as long as they are
    unique.
    \end{changed}

    \subsubsection{Image Encoding} The chromatized image is fed
    through a ViT network that produces a set of image tokens by considering
    the image as a set of non-overlapping patches which are linearly projected
    to the embedding dimension $E$, and then fed through a stack of
    self-attention layers. The ViT network is not pretrained.

    \begin{changed}
    \subsubsection{Sequence Embedding} Each token value $t_i\in\TokenSet{}$ is
    mapped to a learnable embedding vector, along with a learned absolute
    position embedding for each $i$. The two embeddings are summed to form the
    final token embedding.
    \end{changed}

  \subsection{Image-Based Formulation}\label{sec:densepred}

    \def\CNNOut{\mat{Y}}
    \def\CNNOutij{Y_{ij}}
    \def\CNNOccij{O_{ij}}
    \def\CNNMLE{\MLE{\mat{M}}}

    Our baseline is an image-based formulation as in \citeunet{}, where a
    \change{chromatized occupancy grid $\mat{\Chroma{}}(\mat{M})$} is the input
    and the target is a rasterized image of the target line segments. The
    outputs are per-cell occupancy logits $\CNNOutij{}$. At inference time, the
    occupancy state $\CNNOccij{}$ is the maximum likelihood estimate
    \begin{equation} \label{fig:cnnmle}
      \CNNOccij{} = \Iverson{\CNNOutij{}(\change{\mat{\Chroma{}}(\mat{M})}) < 0}
    \end{equation} 
    Marching squares is used to recover the predicted line segments. The loss
    function is per-pixel binary cross-entropy.

\section{Data Synthesis}\label{sec:datagen}
  
  
  \eat{This section details the method used to synthesize training data. Briefly,
  each floor plan is transformed to a set of line segments, and a set of paths
  through the floor plan are generated. Samples are then generated at steps
  along each path as a virtual robot traverses them and updates its occupancy
  grid.}
    Similar to our previous work, we derive training and test data from the
    \ds{}~\cite{aydemir}. Each floor plan is represented as a set of rooms,
    each room being a polygon of line segments categorized as walls, doors or
    windows. For our purposes, only two boolean properties of the segment
    categories are considered: \emph{transparent} and \emph{passable}. Doors
    are assumed to be open and so are both transparent and passable; windows
    are transparent but impassable; and, walls are non-transparent and
    impassable. The segments are collected into two sets, one with impassable
    segments used to generate paths in the floor plan, and one with
    non-transparent segments used for simulating sensor occlusions. Note that
    windows are not only exterior windows, but also glass walls which are
    common in office environments such as the \ds{}.

    Waypoints are first sampled by farthest point
    sampling~\cite{qi2017pointnet++} inside each floor plan. Paths are then
    generated by Dijkstra's algorithm~\cite{dijkstra1959note} between every
    pair of waypoints with a cost based on the truncated distance to the
    nearest wall, so that some minimum wall clearance is maintained where
    possible. A virtual sensor is then simulated along each path at a
    predefined step length, yielding a partial occupancy grid at each step of
    the path built from the scans up to that step by \change{ray marching.
    Incident cells are first marked \F{}, and the terminal cells are then
    marked \O{} or \W{} if the ray is a \emph{hit}, \ie, terminated before the
    sensor's maximum range.} Each occupancy grid is accompanied by its target
    line segments from the floor plan. The target segments are filtered by
    removing subsegments that lie inside \O{} cells.

  \subsection{Axis Alignment}

    The occupancy grid axes are automatically rotated to coincide with the
    visible line segments, which adds invariance in the world-to-robot rotation
    up to an integer multiple of \ang{90}. Alignment also improves fidelity, as
    straight lines in the environment are rotated to coincide with either the
    row or column axis of the occupancy grid which minimizes discretization
    error, which is important for recovering the visible line segments.
    Following~\cite{weiss1994keeping}, a histogram is computed over the angle
    of the line passing through each pair of neighboring LIDAR points, modulo
    \ang{90}. The robot-to-surroundings rotation $\alpha$ is then computed by
    an average of the histogram's mode and the two adjacent bins, weighted by
    frequency. Finally, the LIDAR scans and target line segments are rotated by
    $-\alpha$ to align them to the occupancy grid axes before the occupancy
    grid is rendered.

  \subsection{Subdividing Segments}


    Subdividing segments serves to simplify the prediction problem by allowing
    long segments to be predicted piece by piece. The subdivision algorithm
    must be reproduced by the model function $f_\theta$, so it should be simple
    and \emph{predictable}. Furthermore, increasing the sequence by padding
    tokens improves model performance~\cite{weiss2021thinking,goyal2023think},
    suggesting that shorter segments should improve performance, ceteris
    paribus. In practice, it becomes a compromise between model performance
    versus inference speed and memory usage.

    Our solution is to subdivide the segments by superimposing a
    \numproduct{21x21} regular grid and cutting the segments where they
    intersect this grid. The effect is that from the model's perspective, the
    segments are always subdivided at the same predefined coordinates.


\section{Predicted Information Gain}\label{sec:pig}

  \def\Ma{\ensuremath{\mat{M}}}
  \def\Mb{\ensuremath{\mat{M}'}}
  \def\Maij{\ensuremath{M_{ij}}}
  \def\Mbij{\ensuremath{M'_{ij}}}
  \def\pa{\ensuremath{p_0}}
  \def\pb{\ensuremath{p_1}}
  \def\K{\change{\ensuremath{|\mathcal{C}|}}}

  \def\palebluedot{\protect\colorcircle{mplcyan}}%
  \def\curposdot{\protect\colorcircle{mplC6}}%

  \begin{figure*}[t]
    \def\ii#1{\symbol{40}#1\symbol{41}}%
    \def\igcells{\diagbox{mplC3}{mplC3!60!mplC0}{Information gain cells}}
    \def\sensorsim{\diagbox{mplC0!40}{mplC0!40!mplC3}{Sensor scan}}
    \figcontents{%
      \begin{adjustbox}{clip,trim=2mm,max width=\linewidth}%
        \input{ig_example/ig_example.pgf}%
      \end{adjustbox}
    }%
    \caption{\change{Illustration of how information gain is computed for a
    \palebluedot{}~Frontier location found along \trajline{}~Trajectory. 
    The initial occupancy grid $\Ma$ is shown in \ii{a}, while \ii{b} to \ii{e}
    show the occupancy grid $\Mb$ after a simulated sensor scan in the predicted environment has been
    integrated into $\Ma$, with \igcells{}, \sensorsim{}, and \predsegs{}~Walls.
    Occupancy grid colors as in \cref{fig:murder}. In \ii{b}, walls are
    extracted from \ii{a}, corresponding to the typical way information gain is
    estimated for a frontier in non-predictive autonomous
    exploration~\citerhnbv{}. In \ii{c}, walls are predicted by a model using
    \ii{a} as input to a U-Net predictor as in~\citeunet{}, and in \ii{d} with
    the method proposed in this paper. In \ii{e}, the ground truth walls are
    used. In this example, the naive~$\tilde{I}_n$,
    convolutional~$\tilde{I}_d$, and our approach~$\tilde{I}_f$ differ from the
    ground truth $\GT{I}$ by \qtylist{1593;770;95}{\cells} (relative difference
    \qtylist[list-units = repeat]{122;58.9;7.26}{\percent})
    respectively.}}\label{fig:ig_example}%
  \end{figure*}

  Evaluating generative models is notoriously difficult as there is typically
  no way to quantify the quality of novel generations, and the measure of
  quality is often subjective. A common approach to this issue is to evaluate
  the generation quality by measuring feature statistics from a different
  neural network~\cite{salimans2016improved,pillutla2021mauve}. As argued
  by~\cite{barratt2018note}, generative models are best evaluated by an
  intended application, which in our case is to predict information gain in
  frontier-based autonomous exploration.
    To define information gain, a probabilistic interpretation must be made of
    the occupancy grid \Ma{}. \change{For this definition, independence between cells
    is assumed, and a categorical probability distribution is defined for the
    label of each cell by} \[
    \label{eq:inducedprob}
        \P{\Maij{} = k} = \begin{cases}
             {\K{}}^{-1} & \text{if }\Maij{} = \U \\
             \Iverson{\Maij{} = k} & \text{otherwise}
        \end{cases}
    \]
    \change{If a cell is unknown, its label distribution is uniform, otherwise,
    the distribution is an indicator function of the label, with
    $\Maij{}\in\mathcal{C}$ as defined in \cref{eq:cellstates}.}
    Let $\Mb{}$ denote the occupancy grid $\Ma{}$ after sensor data integration
    at some frontier.
    \change{Information gained in $\Mb$ given $\Ma$} is then defined as
    \[ \label{eq:ig} \change{I(\Mb{} \mid \Ma{})} = \sum_{ij,k} \P{\Mbij{} = k} \log \frac{\P{\Mbij{} = k}}{\P{\Maij{} = k}} \]
    Notice that terms where $\Maij{}=\Mbij{}$, or $\Mbij{}=\U{}$,
    become zero. Assuming known cells in $\Ma{}$ remain the same label, then
    the only non-zero terms will be those where $\Maij{}$ is unknown and
    $\Mbij{}$ is known. In other words, \Cref{eq:ig} is proportional to the
    number of unknown-turned-known cells, \change{which is the definition of
    information gain used in this and other works, \eg, \cite{shrestha2019learned}.}






    Each occupancy grid cell is classified as either a frontier or non-frontier
    based on its 4-connected neighbors. Specifically, a cell is deemed a
    frontier cell if it is free, and amongst its four neighboring cells there
    exists both free and unknown cells. Following this classification, the
    DBSCAN algorithm~\cite{ester1996dbscan} is used to group these frontier
    cells into frontier clusters. The frontier location is the location of the
    cell closest to the average cell location.

\section{Experimental Setup}\label{sec:expsetup}


    Since the \ds{} represents rooms as closed polygons, there are no
    connecting line segments where rooms connect, \eg{}, doorways, making it
    possible to traverse and see inside the walls at these doorways. A
    heuristic is used to insert wall segments in such situations: \change{find
    a bijection from each door segment $uv$ to its nearest door segment $u'v'$
    and insert two wall segments $uu'$ and $vv'$ so that a quadrilateral is
    formed, if $\|uu'\| + \|vv'\| < \qty{2}{\meter}$.} After this, each set of
    segments is canonicalized by first merging \change{nearly identical}
    vertices (within \qty{1}{\mm}), then joining overlapping line segments, and
    finally removing vertices that are not corners; \ie, vertices $v$ with
    exactly two neighbors $u,w$ that form a straight line $uw$ through $v$. 





    \change{The generated paths are filtered by length and number of turns,
    only keeping paths between \qtyrange[range-units = repeat]{5}{100}{\meter}
    long and having at least~3 turns. The virtual sensor returns 720 points per
    scan, and a scan is obtained every \qty{80}{\centi\meter} along the path. A
    cell is \W{} if a hit inside it is within \qty{10}{\mm} from the nearest
    exterior window segment.} A window segment is classified as exterior if
    both its vertices are within some threshold distance (\qty{100}{\mm}) from
    the building perimeter.

    \change{Frontier clusters with less than \qty{3}{\cells} are excluded from
    evaluation, as are those whose center lies within \qty{5}{\cells} of the
    edge of the occupancy grid. Cluster larger than \qty{30}{\cells} are
    divided into subclusters by k-means~\cite{macqueen1967some}.}

    There are on average \num{75} context line segments and
    \num{165} target line segments per sample in \num{6392919} samples from
    \num{164} floor plans. Each occupancy grid represents a
    \qtyproduct{15x15}{\meter} area in \numproduct{121x121}{\cells}. \change{The
    vertex quantization function \cref{eq:quantize} is identical in scale and
    size to the occupancy grid, \ie,
    $\frac{H}{s_y}=\frac{W}{s_x}=\qty{15}{\meter}$ and $H=W=121$. The maximum
    sensor range is $r=\qty{4.5}{\meter}$ unless otherwise stated.}

  \subsection{KTH Dataset Considerations}

    It is important to note that the \ds{} contains duplicated floor plans, and
    that there are strong similarities between floors of a single building. It
    would therefore be an error to simply shuffle the entire dataset, as this
    would contaminate the training set with test data. As in~\cite{iros22}, we
    deduplicate the \ds{} and split it into training and test \emph{before}
    shuffling, and adjust the splitting point such that a single building's
    floor plans are all exclusively training or test data, preventing any cross
    contamination.

  \begin{changed}
  \subsection{Model Hyperparameters}

    In all experiments, the embedding dimension $E=\num{512}$.
    AdamW~\cite{adam} is used with weight decay ${10}^{-2}$, learning rate
    ${10}^{-4}$, and batch size 6. A \pct{10} dropout is also applied. Since
    the dataset is large compared to the model size, the network can be trained
    indefinitely without overfitting; training was stopped when the validation
    loss stopped decreasing. The occupancy grids, visible line segments, and
    target line segments are jittered by one of the eight symmetries of the
    square, \ie, a random combination of mirroring and rotating. Performance
    was not improved by relaxed regularization.

    Top-$p$ sampling~\cite{nucleus} is used for all generation, with
    $p=\pct{80}$. Other choices of $p$ were evaluated but yielded worse
    results. No temperature scaling or repetition penalty is applied.
  \end{changed}




\section{Experimental Results}\label{sec:expresults}

  \def\DMAE#1{$\Delta$MAE \qty{#1}{\cells}}

\begin{figure*}
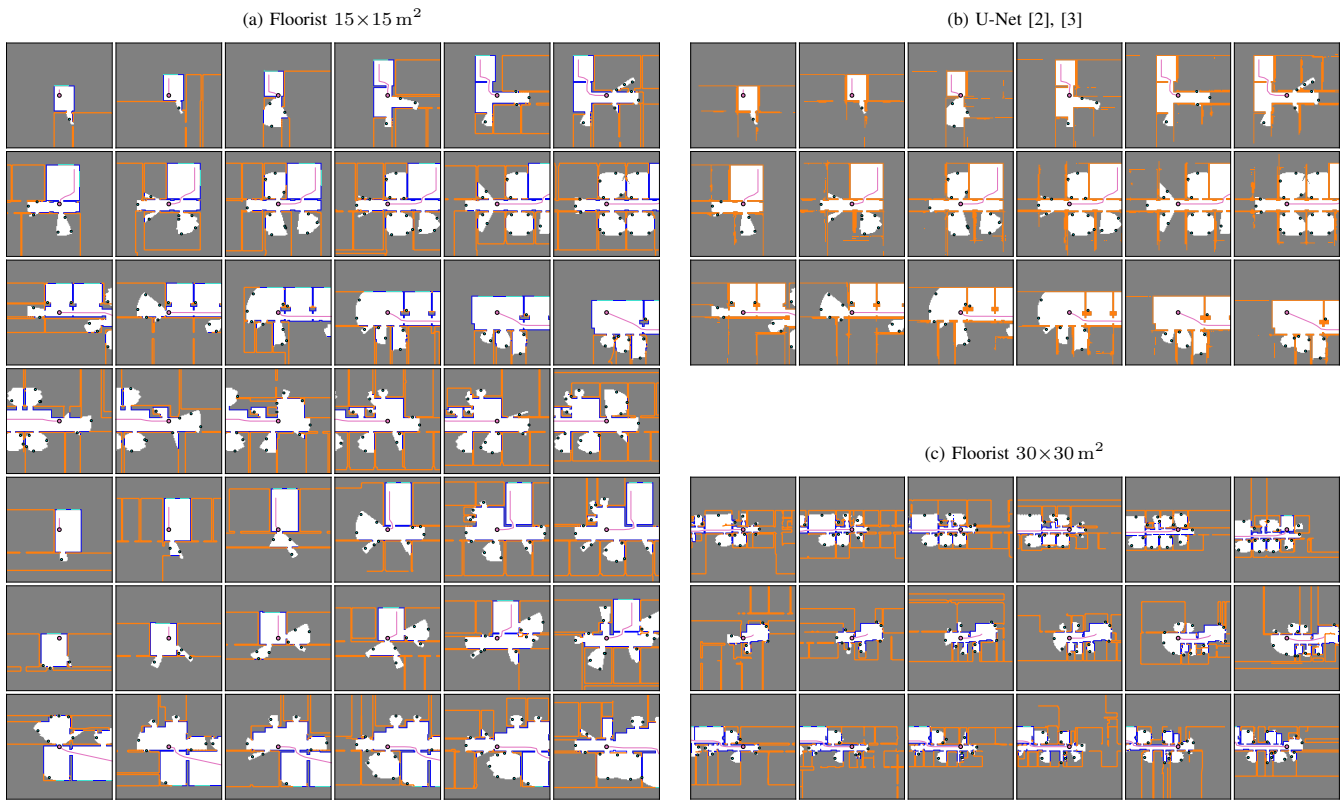
%
      \begin{minipage}[t]{258pt}%
        \scriptsize%
        \centering%
        {(a) Floorist \qtyproduct{15x15}{\m}} \\[4pt]%
        \adjustbox{trim=2mm,clip,max width=250pt}{\input{./florp_compl_15.pgf}}%
      \end{minipage}%
      \begin{minipage}[t]{258pt}%
        \scriptsize%
        \centering%
        {(b) U-Net~\citeunet{}} \\[4pt]%
        \adjustbox{trim=2mm,clip,max width=250pt,scale=0.9935}{\input{./florp_compl_dense.pgf}} \\[9.2145mm]%
        {(c) Floorist \qtyproduct{30x30}{\m}} \\[4pt]
        \adjustbox{trim=2mm,clip,max width=250pt,scale=0.9935}{\input{./florp_compl_30.pgf}}
      \end{minipage}%
      \caption{Examples of predictions on random samples from the test set.
      Each row is a section of a single trajectory, in sequence from left to
      right. \palebluedot{}~Frontier locations used in the information gain
      evaluation are also shown. Other conventions as in \cref{fig:ig_example}.
      Note that in (b), the predicted occupancy grid is shown with
      \legendbox{mplC1}{Occupied cells}. It is advisable to use a digital
      document viewer to zoom the vector graphics.}\label{fig:samples}%
    \end{figure*}

  \begin{changed}
  In \cref{fig:samples}, examples of the generated wall segments are shown for
  six continuous steps of randomly sampled trajectories in the test set.
  Frontier locations used when evaluating predicted information gain are also
  shown. Results from a model with larger area but same network size is also
  shown, showing less coherent output, as the prediction task is harder; \eg,
  there are segments passing through free space, a mistake that the smaller
  model does not tend to make. Finally, results from the image-based predictor
  are also presented, illustrating the difficulty in a pixel-wise approach to
  occupancy regression.
  \end{changed}

  \subsection{Predicted Information Gain}\label{sec:results_ig}

    Predicted information gain is evaluated using the following different sets
    of wall segments, illustrated in \cref{fig:ig_example}:

    \def\convnetcite{\cite{ronneberger2015unet,liu2022convnext,dosovitskiy2020vit,he2016resnet,chen2017deeplab,zwecher2022integrating}}



    \subsubsection{Naive} Assume that only what has been observed to be
    occupied is occupied, \ie{}, no segments are predicted and only the walls
    visible in the occupancy grid occlude the sensor. This is a common approach
    in non-predictive exploration planning~\citerhnbv{}, and is equivalent to
    assuming that unknown space can be considered sensor transparent.

    \subsubsection{U-Net}
    {U-Net}~\cite{ronneberger2015unet} is a fully-convolutional image
    segmentation network, chosen as a baseline as it has been used in previous
    work on occupancy grid prediction for autonomous exploration~\citeunet{}.
    The network was not pretrained.
    Several network architectures were evaluated for image-based
    prediction~\convnetcite{}. All architectures reached convergence at the
    same loss value, suggesting that convergence is caused by the intrinsic
    difficulty of image-based prediction.

    \begin{changed}
    \subsubsection{Floorist} A \numproduct{121x121} vertex grid
    with $L=6$ attention layers, 8-way MHA, \num{4096} GeLU units, and
    $E=\num{512}$ embedding dimensions. The image encoder is a three-layer
    ViT~\cite{dosovitskiy2020vit} with patch size 6 (\ie, $P=400$), 8-way MHA,
    and \num{4096} hidden units.
    \end{changed}

    \subsubsection{Ground Truth} Using the actual walls of building, \ie, true information gain.




    \begin{changed}
    \begin{figure}[tb]
        {%
          \vspace{0.8mm}%
          \vspace{0.8mm}%
          \begin{adjustbox}{clip,trim=2mm,width=\linewidth}%
            \hspace*{\fill}%
            \input{cdf.pgf}%
            \hspace*{4mm}%
            \hspace*{\fill}%
          \end{adjustbox}%
        }%
        \caption{\change{Cumulative distribution function $F$ of absolute
        error $|d| = |\Est{I} - \GT{I}|$ in predicted information gain $\Est{I}$
        from the true information gain $\GT{I}$ using line segments from
        \colorline{mplC0}~Naive, \colorline{mplC1}~{U-Net}, and
        \colorline{mplC2}~{Floorist}. $N=\num{1464140}$.}}\label{fig:cdf}
    \end{figure}

    %


    In \cref{fig:cdf}, the cumulative distribution function of absolute errors
    in predicted information gain is presented for the three evaluation targets
    on test data. Image-based prediction provides substantial improvements over
    naive non-predictive estimation, and our sequence-based approach in turn
    provides substantial improvements over image-based prediction.
    \pct{95} confidence intervals for the median absolute error (MAE) are
    \numlist{1195\pm2;452\pm1;236\pm1} cells. The intervals are computed by
    bootstrapping with \num{1000} trials.
    A \emph{two-sample Kolmogorov-Smirnoff test} (KS) indicates whether a pair
    of CDFs differ significantly by measuring the largest vertical gap.
    Image-based predictions are significantly more accurate than naive
    estimation (\DMAE{745}, KS \pct{34.5}), and Floorist predictions are in
    turn significantly more accurate than image-based prediction (\DMAE{217},
    KS \pct{16.2}). \end{changed}

  \begin{changed}

  \subsection{Scale and Sensor Range}

    To assess the impact of scale, the information gain evaluation is performed
    on a model with twice the area, \qtyproduct{30x30}{\meter}. The occupancy grid size is limited by GPU memory, and
    remains \numproduct{121x121}~cells. The resolution is
    therefore four times lower (\qty{16.27}{\cells\per\square\meter} vs
    \qty{65.07}{\cells\per\square\meter}). The larger area resulted in \pct{75}
    longer token sequences, requiring significantly more GPU memory; batch size
    was therefore reduced to 2. Information gain is evaluated as in
    \cref{sec:pig}, with two sensor range settings: $r=\qty{4.5}{\meter}$ and
    $r=\qty{9}{\meter}$. To ensure the results are comparable, information gain
    is computed at the base resolution. The MAE of each model at each setting
    is reported in \cref{tab:scaleeval}, with the frequency of overestimating
    and underestimating of the information gain. The first three rows are the
    results reported in \cref{sec:results_ig}. The performance of the
    larger-scale model is slightly worse than the base model in the shorter
    range setting (\DMAE{40}). The large-scale model is significantly better
    than the base model in the 9-meter case (\DMAE{679}), because the longer
    sensor range often reaches outside the \qtyproduct{15x15}{\m} prediction
    area of the base model, giving an overestimating effect similar to naive
    estimation. This is reflected in the sharp increase in frequency of
    overestimation. 





  \end{changed}

  \def\Err{d}%
  \def\tr#1&#2&#3&#4&#5\\{#1 & #2 & \phantom{ 1195 }\llap{#3} & \phantom{ 0.00 }\llap{#4} & \phantom{ 00.0 }\llap{#5} \\}

  \begin{table}
    \caption{Evaluation of Scale and Sensor Range}\label{tab:scaleeval}
    \centering
    \begin{tabular}{ccccc}
      \toprule
      Range  (\meter) & Area                        & Median (\cells)     & Under (\%)                             & Over (\%)                             \\
      $r$             & (\unit{\m\squared})         &  $|\Est{I}-\GT{I}|$ & $\langle[\Est{I}<\GT{I}]\rangle$ & $\langle[\Est{I}>\GT{I}]\rangle$ \\
      \midrule%
      \tr \num{4.5}  & Naive                        & 1195            &     {0.00} &     {98.0} \\
      \tr            & U-Net                        & 452             &     {11.3} &     {85.9} \\

      \tr            & \numproduct{15x15}       & \bfseries 236   &     {28.5} &     {66.5} \\
      \tr            & \numproduct{30x30}       & 276             &     {34.0} &     {58.7} \\
      \midrule%
      \tr \num{9.0}  & \numproduct{15x15}       & 1475            &     {19.0} &     {80.1} \\
      \tr            & \numproduct{30x30}       & \bfseries 796   &     {31.5} &     {66.0} \\
      \bottomrule
    \end{tabular}
  \end{table}

  \subsection{Model Ablations}

    \def\tr#1&#2&#3\\{#1 & \rlap{#2}\phantom{ 1.23 (+0.123) } & \rlap{#3}\phantom{ 12.3 (-1.23) } \\}
    \def\Accuracy{a}
    \def\FlooristBase{Floorist~$15\tymes{}15 \unit{\m\squared}$}
    \def\FlooristBase{Floorist~\qtyproduct{15x15}{\m}}

    \begin{table}[tb]
      \centering%
      \begin{threeparttable}
        \caption{Evaluation of Model Ablations on Test Set}\label{tab:ablations}
        \begin{tabular}{lcc}
          \toprule
                  Ablation \hspace{25mm}& Loss (bits)   & Accuracy (\%) \\
                                        & $\Expect{L}$  & $\langle[\GT{t} = \MLE{t}]\rangle$ \\
          \midrule
          \tr \FlooristBase{}           & 1.13          & 81.7         \\
          \tr No Window Labels          & 1.17 (+0.043) & 81.3 (-0.38) \\
          \tr Shared Vertex Embedding   & 1.20 (+0.065) & 80.8 (-0.87) \\
          \tr No Context Segments       & 1.23 (+0.103) & 80.1 (-1.55) \\
          \tr No \change{Image Encoder} & 1.40 (+0.267) & 78.1 (-3.54) \\
          \bottomrule
        \end{tabular}
        Values in parentheses are relative to the first row. $L$ is defined as
        in \cref{eq:loss}, and $\MLE{t}$ is the maximum likelihood estimate.
      \end{threeparttable}%
    \end{table}

    We ablate some key components of our model and evaluate the impact on its
    performance on the test set in terms of the loss and the accuracy of the
    maximum likelihood estimate.
     The following ablations are
    reported in \cref{tab:ablations}: \W{} labels replaced by \O{} in training
    and test occupancy grids; embedding weights of visible segments is shared
    with embedding weights of the target segments; without visible segments
    \change{as cross-attended tokens, \ie, only using the occupancy grid as
    context; and, without the image encoder, \ie, only using visible segments
    as context. We see that each ablation degrades performance, and that the
    occupancy grid context is most important. This may be because the occupancy
    grid is the only volumetric representation of the environment, and
    indicates where free versus unknown space is.}

    \change{Window labels have the smallest impact on the reported metrics} since
    windows are relatively uncommon in the dataset. Window labels were included
    as a cue for the building exterior, so the effect of those labels should be
    evaluated by evaluating if they prevent the model from predicting walls
    outside the building exterior. We specifically look at instances where at
    least one occupancy grid cell has the \W{} label, and compute the length of
    predicted line segments that fall outside the building perimeter as a
    proportion of the total predicted segment length. Window labels reduced the
    length of segments outside the building perimeter by \pct{22.4}, from
    \pct{10.93} to \pct{8.48}, indicating that the model does respond to the
    window cues.

\begin{changed}
\subsection{Reconstructed Floor Plans}\label{sec:realworld}
\end{changed}

  The software library \emph{RoomPlan}~\cite{roomplan} enables online
  reconstruction of floor plans from real-world sensor data in cluttered
  environments. As a \change{test} of the generality of our approach, we show
  that our method can use such reconstructions. RoomPlan runs on a modern
  smartphone using an image sensor and a solid-state LIDAR sensor, and produces
  a parametric 3D model of the environment as a set of objects of some
  predefined types: walls, doors, windows, and miscellaneous furniture. From
  this model, 2D line segments are derived by the orthogonal projection of the
  walls and windows onto the ground plane. A LIDAR sensor is simulated along
  the trajectory, and the floor plan is predicted from the resulting occupancy
  grid. The reconstructed 2D floor plan, the trajectory, and a wall prediction
  is shown \cref{fig:roomplan}. \change{The model correctly infers that there
  is a corridor outside, and that there is an adjoining room; however, it
  predicts the room to be in the middle of the corridor, not the end, as is the
  actual case.}
  %

  \newlength{\sx}
  \setlength{\sx}{0.3455723542\linewidth}
  \newlength{\sy}
  \setlength{\sy}{0.3088552916\linewidth}
  \begin{figure}[bt]%
    {%
      \def\frame#1{#1}%
      \footnotesize%
      \frame{\includegraphics[trim=2mm 2mm 2mm 2mm,clip,width=\sx]{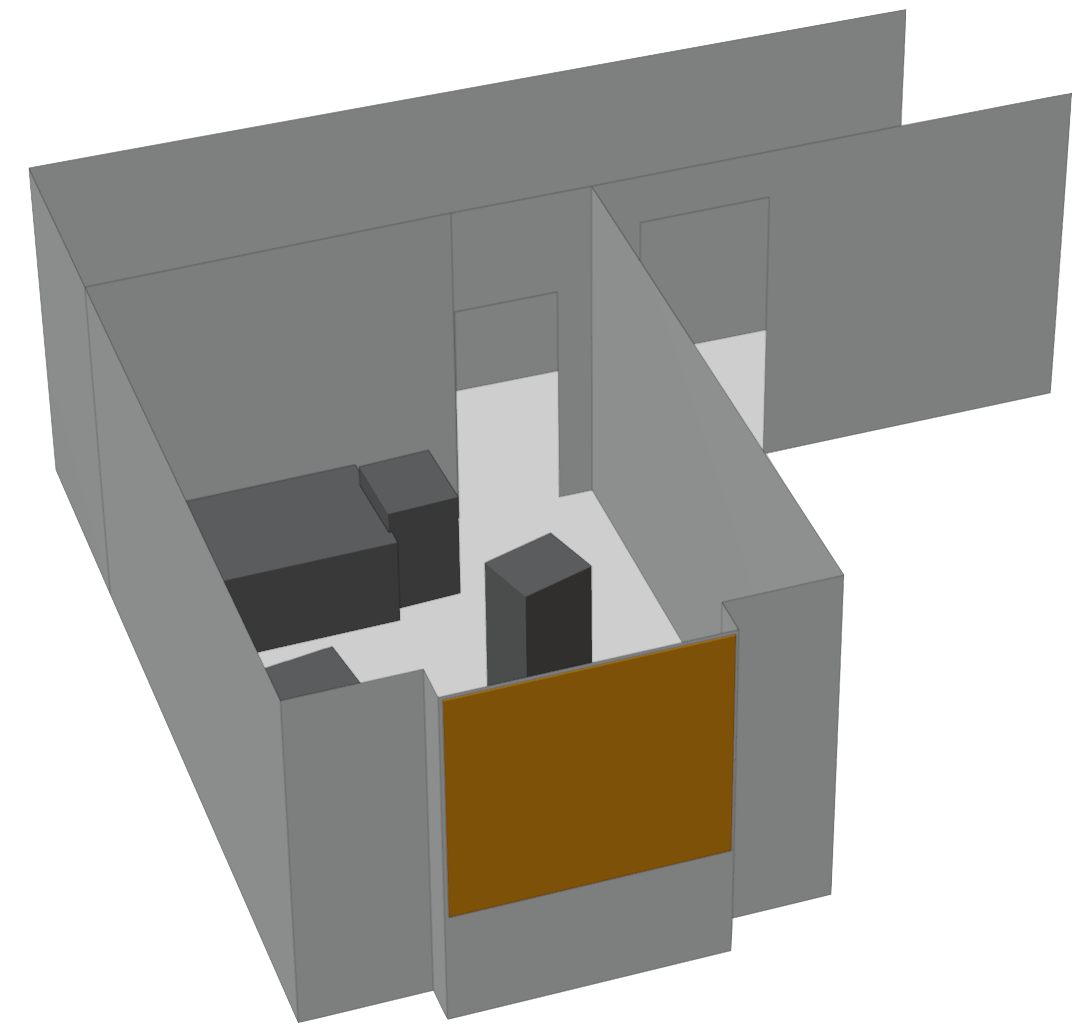}}%
      \hfill%
      \frame{\begin{adjustbox}{trim=2mm 2mm 2mm 2mm,clip,height=\sy}\input{roomplan/lidarsim.pgf}\end{adjustbox}}%
      \hfill%
      \frame{\begin{adjustbox}{trim=2mm 2mm 2mm 2mm,clip,width=\sx}\input{roomplan/sample_traj.pgf}\end{adjustbox}}%
      \\%
      \frame{\makebox[\sx]{(a)}}\hfill\frame{\makebox[\sy]{\hspace*{2mm}(b)}}\hfill\frame{\makebox[\sx]{\hspace*{3mm}(c)}}%
      %
    }%
    \caption{Illustration of (a)~a reconstructed floor plan
    from~\cite{roomplan}, with walls, a window, and furniture; (b)~simulated
    \trajline{}~Trajectory inside the floor plan derived from (a) with sensor
    scans; and (c)~the axis-aligned occupancy grid, and an example of
    \predsegs{}~Predicted walls from Floorist.}%
    \label{fig:roomplan}%
  \end{figure}

\vspace{5mm}
\section{Conclusion}

  In this work, we have presented an attention-based generative model for floor
  plans conditioned on realistic sensor data. We have shown that our model can
  cope with the high dimensionality of a time-dependent occupancy grid
  representation with a realistic sensor model and make competent predictions
  as quantified by evaluating the predicted information gain. The approach
  offers advantages over traditional image-based predictions at the cost of
  longer inference time, though autoregressive sampling performance is an
  active area of research. We have also shown that the approach is robust
  enough to be used in real-world environments, demonstrated by using an
  off-the-shelf floor plan mapping solution as the source of floor plan data.
  In the future, we aim to extend our real-world demonstration to create a
  real-time floor plan prediction system from sensor data. Another interesting
  direction is adversarial multi-agent contexts where intuiting the surrounding
  is important, such as pursuit-evasion and other search games.


  \balance
  \bibliographystyle{IEEEtrandiy}
  \bibliography{your}


\end{document}